%% file: root.tex
\documentclass[letterpaper,10pt,conference]{ieeeconf}
\usepackage[utf8]{inputenc}
\usepackage{times}
\usepackage{epsfig}
\usepackage{graphicx}
\usepackage{amsmath}
\usepackage{amssymb}
\usepackage{booktabs}
\usepackage{epsfig}
\usepackage{etoolbox}
\usepackage{floatrow}
\usepackage{graphicx}
\usepackage{lipsum}
\usepackage{microtype}
\usepackage{nicefrac}
\usepackage{url}
\usepackage{multirow}
\usepackage{cite}
\usepackage{cleveref}
\usepackage{multirow}

\usepackage{floatrow}
\usepackage{colortbl}

\usepackage[table]{xcolor}
\definecolor{red}{rgb}{0.95,0.4,0.4}
\definecolor{green}{rgb}{0.55,1.0,0.55}
\definecolor{lightgreen}{rgb}{0.75,1.0,0.75}
\definecolor{blue}{rgb}{0.4,0.4,0.95}
\definecolor{darkblue}{rgb}{0,0,0.8}
\definecolor{darkred}{rgb}{0.8,0,0}
\definecolor{darkgreen}{rgb}{0,0.5,0}
\definecolor{grey}{rgb}{0.6,0.6,0.6}

\definecolor{col1}{RGB}{232, 161, 148}
\definecolor{col2}{RGB}{148, 187, 232}

\usepackage{stfloats}
\usepackage{pgfplots}

\usepackage[backgroundcolor=white,linecolor=red,bordercolor=none]{todonotes}

\crefname{table}{Tab.}{Tab.}
\Crefname{table}{Tab.}{Tab.}
\crefname{figure}{Fig.}{Fig.}
\Crefname{figure}{Fig.}{Fig.}

\IEEEoverridecommandlockouts

\title{Monocular Depth Estimation with Self-supervised Instance Adaptation}
\author{
Robert McCraith\and Lukas Neumann\and Andrew Zisserman \and Andrea Vedaldi
\thanks{Authors are with Visual Geometry Group, Dept.~of Engineering Science, University of Oxford {\tt\scriptsize \{robert,lukas,az,vedaldi\}@robots.ox.ac.uk}}
}

\begin{document}
\maketitle
\thispagestyle{empty}
\pagestyle{empty}
\begin{abstract}
Recent advances in self-supervised learning have demonstrated that it is possible to learn accurate monocular depth reconstruction from raw video data, without using any 3D ground truth for supervision.
However, in robotics applications, multiple views of a scene may or may not be available, depending on the actions of the robot, switching between monocular and multi-view reconstruction.
To address this mixed setting, we proposed a new approach that extends any off-the-shelf self-supervised monocular depth reconstruction system to use more than one image at test time.
Our method builds on a standard prior learned to perform monocular reconstruction, but uses self-supervision at test time to further improve the reconstruction accuracy when multiple images are available.
When used to update the correct components of the model, this approach is highly-effective.
On the standard KITTI benchmark, our self-supervised method consistently outperforms all the previous methods with an average 25\% reduction in absolute error for the three common setups (monocular, stereo and monocular+stereo), and comes very close in accuracy when compared to the fully-supervised state-of-the-art methods.


\end{abstract}

\input{intro}

\input{previouswork}
\input{method}

\input{experiments}
\input{conclusion}
\bibliographystyle{IEEEtran}
\bibliography{bibliography}
\nocite{joo2020exemplar}
\end{document}

%% file: intro.tex
\section{Introduction}\label{s:intro}
\newcommand{\turnheightnew}{0.2\columnwidth}

\begin{figure*}[b]
    \centering
        \begin{tabular}{@{\hskip 1mm}c@{\hskip 1mm}c@{\hskip 1mm}c@{\hskip 1mm}c@{\hskip 1mm}c@{}}

        {\rotatebox{90}{\hspace{2mm}\small input}} & \includegraphics[width=\turnheightnew]{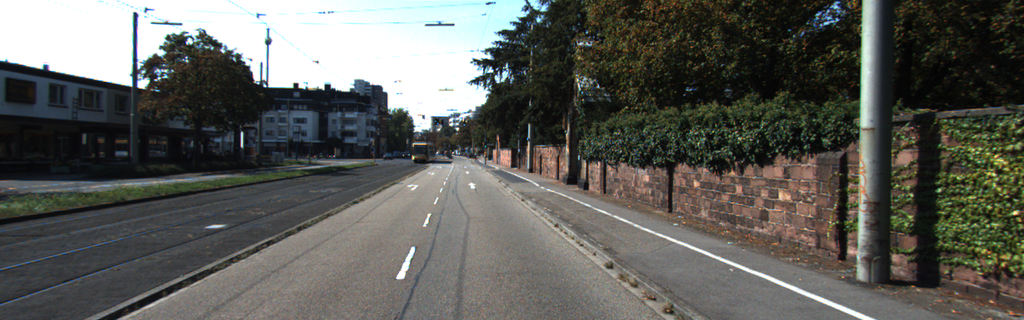} & \includegraphics[width=\turnheightnew]{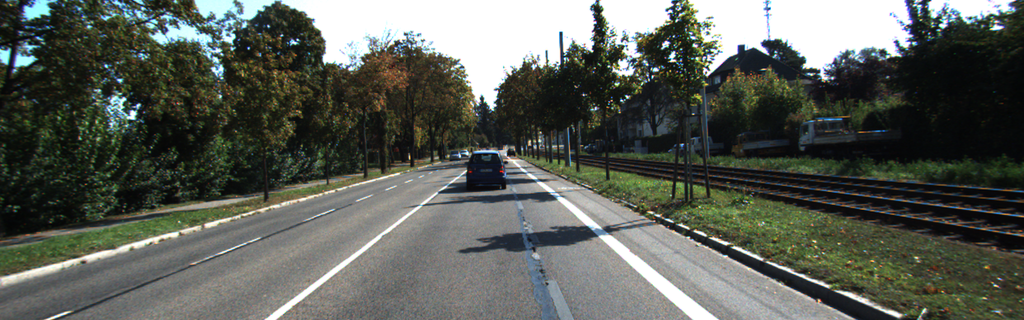} & \includegraphics[width=\turnheightnew]{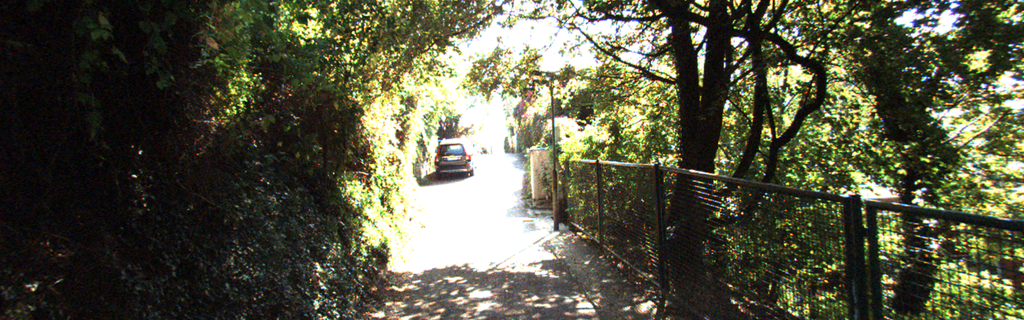} & \includegraphics[width=\turnheightnew]{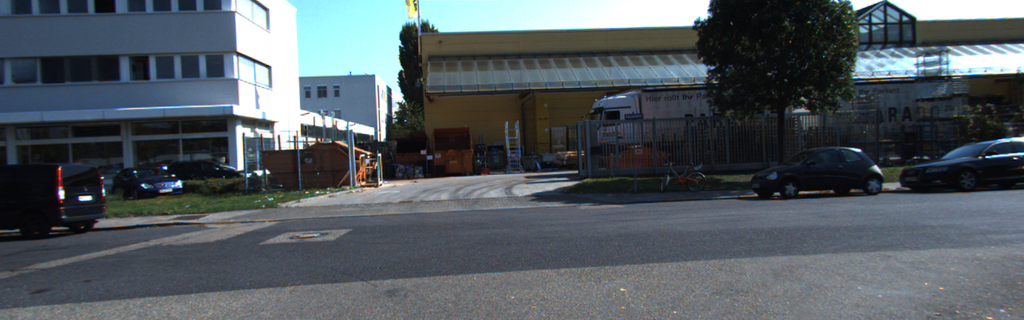} \\

        {\rotatebox{90}{\hspace{0mm}\small baseline}} & \includegraphics[width=\turnheightnew]{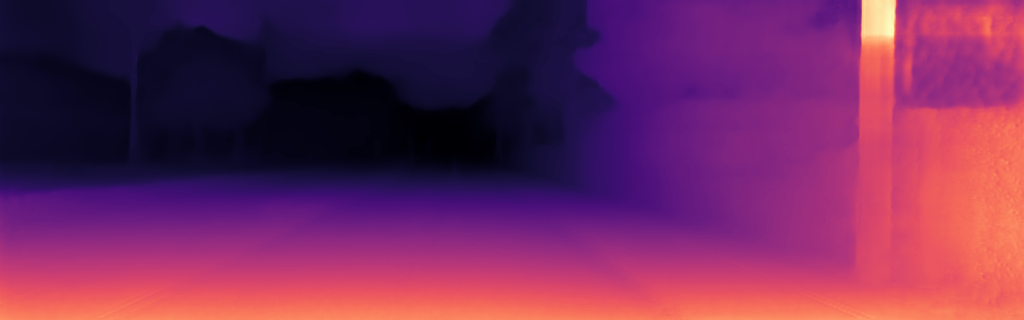} & \includegraphics[width=\turnheightnew]{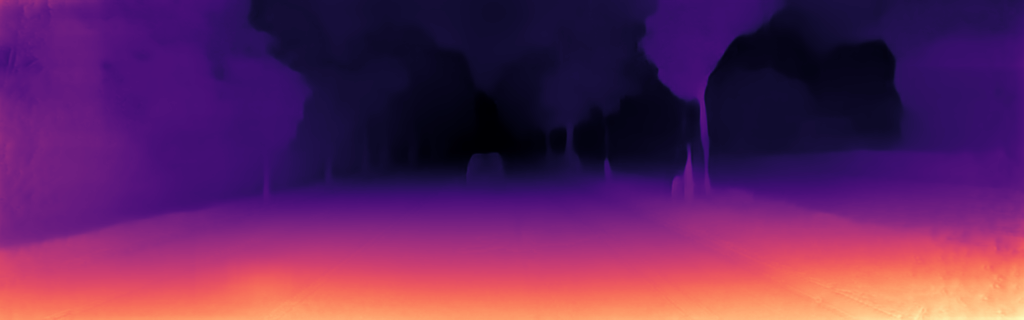} & \includegraphics[width=\turnheightnew]{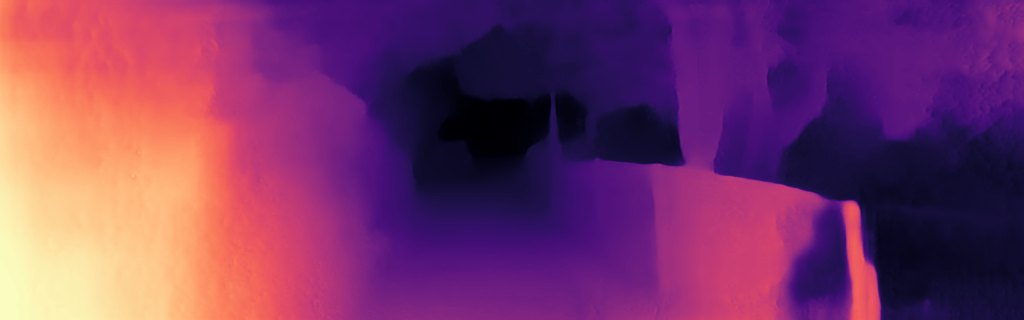} & \includegraphics[width=\turnheightnew]{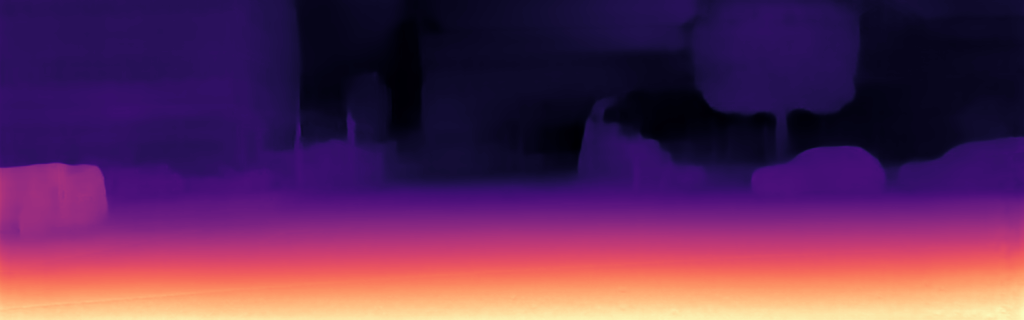} \\

        {\rotatebox{90}{\hspace{2mm}\small ours}} & \includegraphics[width=\turnheightnew]{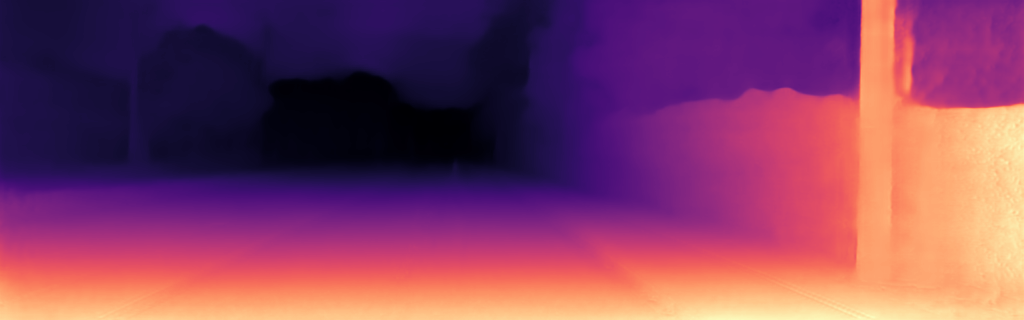} & \includegraphics[width=\turnheightnew]{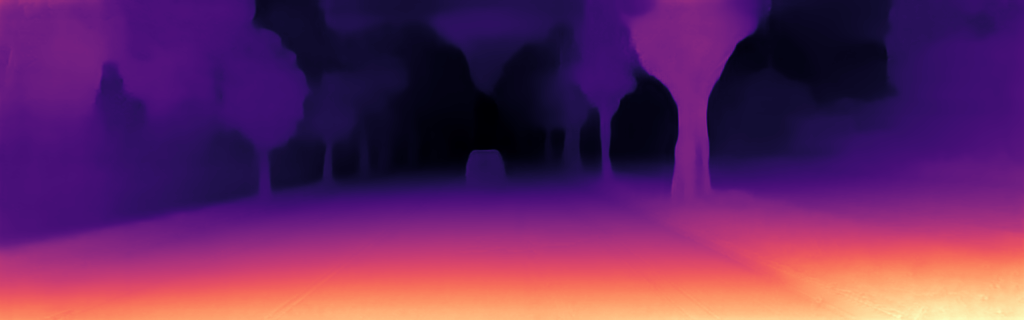} & \includegraphics[width=\turnheightnew]{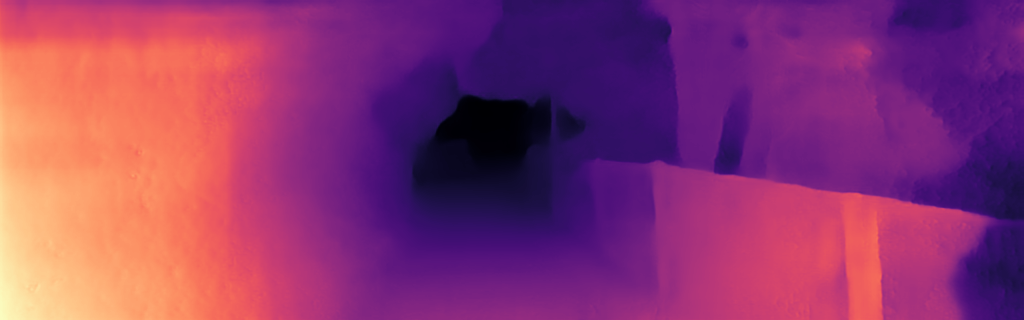} & \includegraphics[width=\turnheightnew]{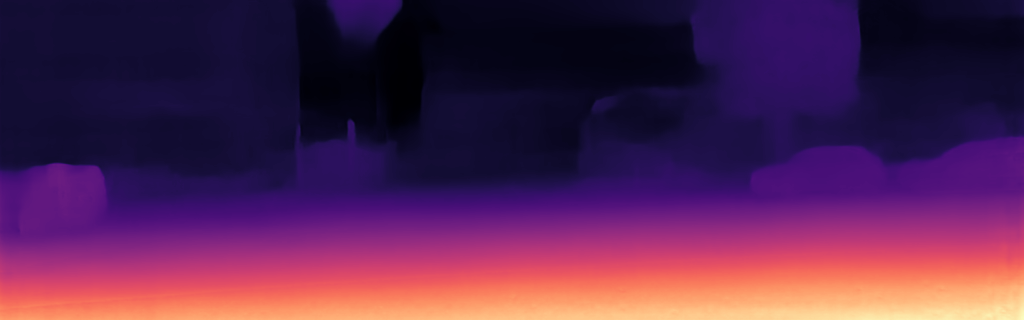} \\

        {\rotatebox{90}{\hspace{0mm}\small LiDAR}} & \includegraphics[width=\turnheightnew]{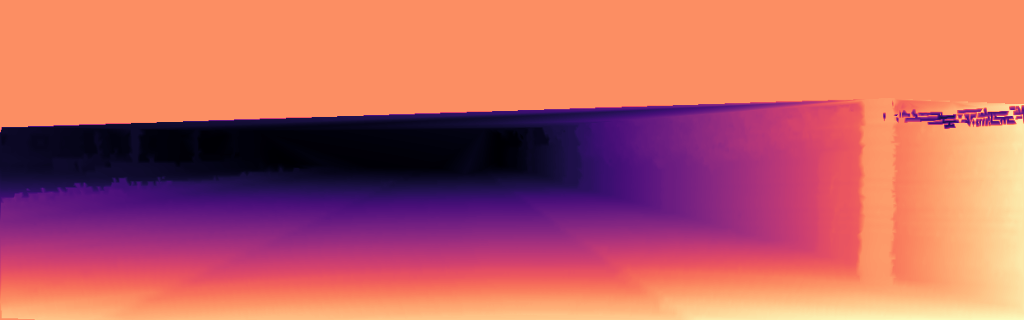} & \includegraphics[width=\turnheightnew]{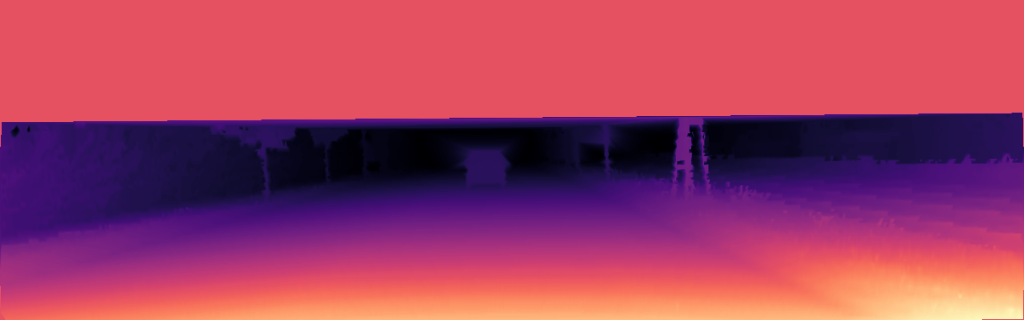} & \includegraphics[width=\turnheightnew]{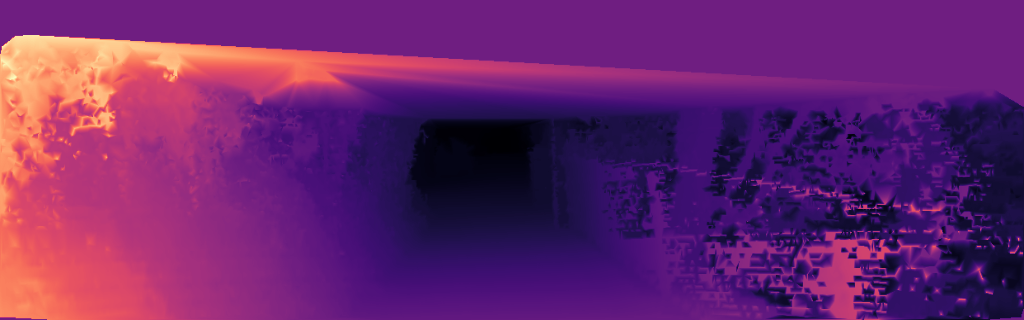} & \includegraphics[width=\turnheightnew]{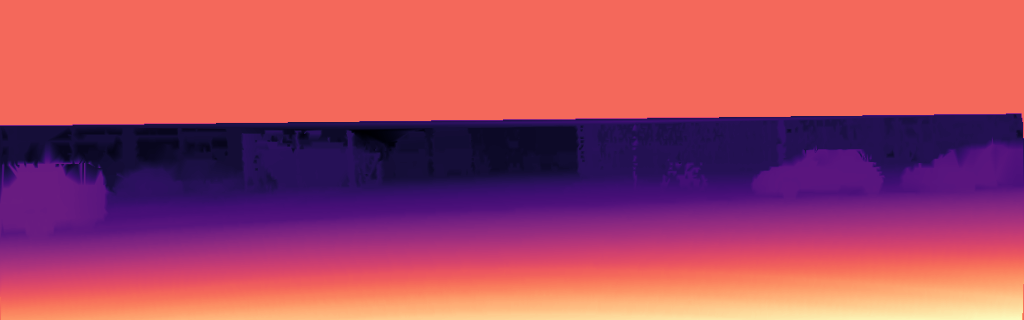} \\

        \end{tabular}
        \caption{Qualitative samples from the KITTI dataset\cite{Geiger2012CVPR}. Self-supervised instance adaptation during inference consistently improves depth estimation across different challenging scenarios. From top to bottom: input image, the baseline state-of-the-art method\cite{monodepthv2}, our method and the interpolated LiDAR ground truth. \vspace{-20pt}}
        \vspace{-10pt}
\end{figure*}

Using images to reconstruct the 3D geometry of an environment, for example in the form of a depth map, has important applicators in robotics, autonomous driving, augmented reality, scene understanding and more~\cite{poggi2018towards,zhou2017,Fu_2018_CVPR,yang2019inferring,8593864}.
Since inferring depth from a single image is ill-posed, traditional implementations use multiple views, obtained either from a moving camera or from a multi-camera setup (usually a stereo pair).
Yet, as humans, we are able to navigate an environment or even drive with a single eye~\cite{hochberg1952familiar}.
In general, we can perform \emph{monocular depth prediction}, estimating depth from a single 2D image.

Monocular depth prediction cannot be obtained via visual geometry alone.
Instead, the problem can only be solved by using a powerful prior on the possible 3D shape and appearance of objects and scenes.
Such a prior must necessarily be \emph{learned from data}, which explains why humans are good at this task.
However, with modern deep learning techniques, it is now possible to learn effective 3D priors for use in algorithms.
Even more remarkably, recent \emph{self-supervised} approaches~\cite{monodepth17,monodepthv2,zhou2017,mancini2016fast} have shown that 3D priors can be learned with no access to 3D ground-truth nor multiple camera setups, using only \emph{videos shot from a single camera}.
This avoids ad-hoc setups involving additional modalities such as LiDARs to collect ground-truth 3D data for training.

Note that, since robots are \emph{supposed} to move, they can usually capture images of the scene from multiple viewpoints, so it may not be obvious why monocular depth estimation is useful in this context.
An answer is that the robot often needs to estimate geometry  \emph{before} moving.
Furthermore,  monocular depth estimation is indifferent to other objects moving in the scene, which breaks the rigidity assumption of multi-view geometry.
Still, there is a clear and unexplored opportunity in developing new methods that can work in a \emph{mixed setting}, combining the benefits of monocular and multi-view depth estimation, depending on which data becomes available to the robot.

In this paper, we observe that, since self-supervised monocular depth estimation learns from the consistency between subsequent video frames~\cite{zhou2017} with no need for any external ground-truth signal, the \textit{same principle can be exploited during inference} in addition to just during training.
We show that, given an existing monocular depth prediction model and a careful specification of which components to optimize, we can update the relevant model parameters (weights) at inference time through standard back-propagation. We also show that our method can seamlessly be extended to mixed monocular and multi-view depth estimation.

This idea allows to use the strong monocular depth prediction prior learned by the model during training to robustly interpret multiple views at test time, improving individual depth predictions.
Furthermore, this refinement process can also be applied \emph{continually} during inference, which not only improves the run-time of the method but also is a form of domain self-adaptation: for example, in a driving scenario, this allows a network trained on a dataset representative of one location to be deployed in areas which  look different, automatically fine-tuning the model.

We show empirically that our depth estimation method significantly outperforms all other self-supervised depth estimation baselines, without requiring any re-training, and very nearly matches the quality of fully-supervised systems. We also significantly outperform recent methods which observe and aggregate information from numerous frames at test time\cite{patil2020dont}.

We also believe this idea could be further extended in several ways.
First, application of our method is optional and can be engaged only when motion is observed, reverting back to standard monocular depth estimation otherwise.
Second, we show that it can benefit from multiple sensors such as stereo pairs that have also been used for self-supervision~\cite{monodepth17}.
Third, it can likely be extended to many other scenarios where self-supervision is applicable, to improve inference or perform self-adaptation at test time.

%% file: previouswork.tex
\section{Previous Work}

\subsection{Other Sensors}
From practical perspective, depth in a scene can be estimated using additional sensors such as LiDARs or radars, but these modalities have their own limitations: LiDARs have problems in certain weather conditions~\cite{Bijelic_2018}, they only produce sparse depth maps which may not be enough for far-away or small objects and they have lower refresh rate when compared to many other sensors (10--15Hz~\cite{velodyne64}).
Similarly, radars suffer from reflections and interference and they struggle to detect small or slowly-moving objects~\cite{radar-shortcomings}.
Stereo vision systems are very sensitive to precise calibration and could face problems with misalignment of cameras following even a negligible collision, but the main drawback is the lack of redundancy --- a moving autonomous vehicle or a robot in urban environment cannot simply stop working if for example one of the cameras becomes occluded by dirt, as that would be potentially very unsafe.
By uniting monocular and multi-view reconstructions setups in a robust manner, our method can help to address these challenges.

\subsection{Supervised Depth}
Owing to the scale ambiguity present when capturing images with a single camera the prediction of depth from a single image is ill-posed as the same image can result from scenes at various scales.
Early approaches to this problem revolved around the use of geometry to extract point to point correspondences between images and triangulating to estimate depth\cite{FlynnNPS15, scharstein2002a}.
Through the use of deep learning solutions to this problem have emerged which utilise additional inputs for supervision typically in the form of LiDAR, RGB-D cameras and stereo pairs. This allows a model to be trained to exploit the relationship between colour images and their corresponding depth.

Eigen et al.\cite{EigenF14} applied multi-scale networks to refine estimated depth maps from low spatial resolution to high resolution depth maps independently.
Xie et al.\cite{XieGF16} used skip-connections to fuse low resolution information dense layers with higher resolution information sparse earlier layers to refine predictions further. {}\cite{LainaRBTN16,GargBR16,KuznietsovSL17} apply multi-layer deconvolutional networks to recover coarse-to-fine depth while others rely on conditional random fields to improve fine details~\cite{Wang_2015_CVPR, LiuSLR15}.
DORN~\cite{Fu_2018_CVPR} introduced space-increasing discretization to transform depth prediction from a regression problem to one of ordinal regression, encouraging faster convergence and improving accuracy. Lee et al.\cite{Lee19} utilise local planar guidance layers at multiple stages in the decoder to improve performance.

\subsection{Self-supervised Depth}
Even with the use of additional sensors capturing ground truth data is not without difficulty. LiDAR produces sparse point clouds, using RGB-D cameras in bright environments results in unstable depth maps and stereo setups require precise calibration and lose accuracy quickly as distance increases. With these problems in mind there has been a recent interest in unsupervised methods for depth map prediction. Early works in this area focused on using stereo images in training such as~\cite{XieGF16} which proposed prediction of discretized depth for novel view synthesis for VR and 3D video applications, {}\cite{GargBR16} extended this approach to continuous values and Monodepth~\cite{monodepth17} added a left-right depth consistency term to the loss to further improve results.
This concept was generalized to remove the need for stereo pairs and instead predict the pose between two sequential frames in a video during training as seen in SfMLearner~\cite{zhou2017}.
Unlike the stereo version where a scene is captured from two viewpoints simultaneously with a sequence of frames, non-stationary objects can move causing problems when comparing an image to another warped to have the same viewpoint, resulting in many methods employing another output mask to discount these regions. {}\cite{Casser19, Yin18} decompose the image into rigid and non-rigid components using flow or semantic maps to reduce this re-projection error.
In Monodepth~V2~\cite{monodepthv2} this problem is tackled using a per pixel minimum re-projection loss which aims to exclude pixels that have become occluded frames surrounding the current frame which would normally results in large error which effects the average error computed for this location.

%% file: method.tex
\section{Method}\label{s:method}

\begin{figure}[b]
    \centering
    \includegraphics[width=\textwidth]{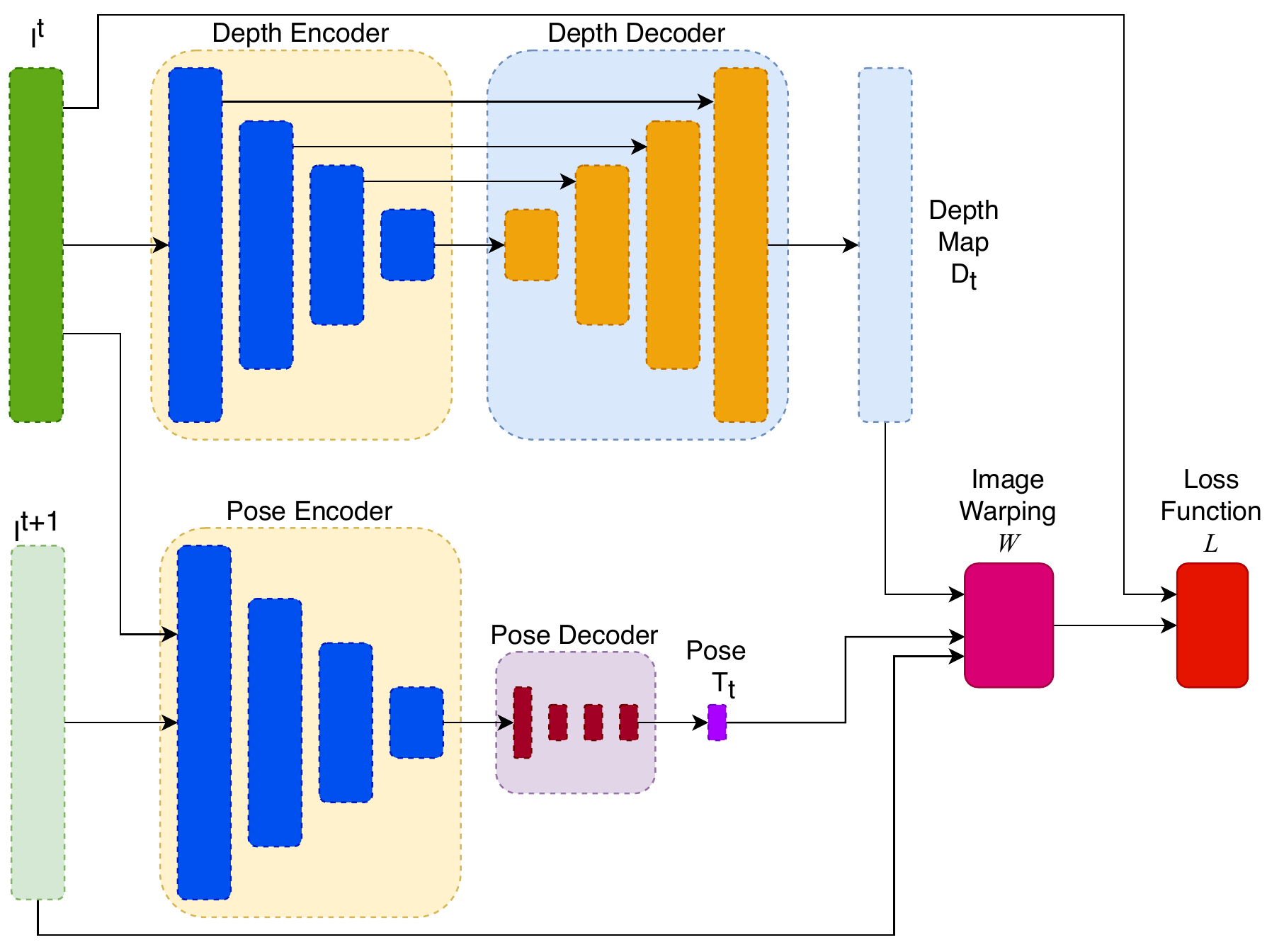}
    \caption{Monocular depth estimation pipeline\cite{monodepthv2}. Modules enabled for instance adaptation during inference marked with  yellow background
    \vspace{-15pt}
    }\label{fig:architecture}
    \vspace{-10pt}
\end{figure}

\input{tables/interventions}

In this section, we describe the architecture of our network and the self-supervised loss function used in training and in inference. We then show how to update certain parameters (weights) of a trained model dynamically at run time for each image instance to optimize depth prediction for each specific image instance individually.

\subsection{Self-supervised Depth Estimation}
In line with monocular depth estimation literature\cite{GargBR16, monodepth17, monodepthv2}, we formulate the depth estimation problem as novel view synthesis, where a network $\Psi$ produces two auxiliary variables --- a dense depth map $D$ and a pose estimate $T$.
Given a pair of input images (two subsequent frames in a video, left \& right images from a stereo camera, etc.), the auxiliary variables $D, T$ are then used to warp one of the images into another, enforcing visual consistency between the warped and the other image.

More formally, and assuming two sequential images $I^t$ and $I^{t+1}$ as an input, the network $\Phi$ is trained to produce estimates $D_t$ and $T_t$ to minimize the appearance loss between the original $I^t$ and the synthesized image $\hat{I}^t = \mathcal{W}(I^{t+1}, D_t, T_t, K)$, where $\mathcal{W}$ is the warping operation\cite{JaderbergSZK15} and $K$ is the camera intrinsics assumed to be known.
We thus have:

\begin{align}
\mathcal{L}(I^t, I^{t+1}) &= \alpha E_{\mbox{\scriptsize p}}(I^t, \hat{I}^{t}) + E_{\mbox{\scriptsize dis}}(I^t, \hat{I}^{t}) \label{eq:lossFunction}\\
\hat{I}^t &= \mathcal{W}(I^{t+1}; D_t, T_t, K) \\
D_t, T_t &= \Psi(I^t, I^{t+1})
\end{align}
where, for training, the network $\Psi$ is optimized using the loss function~\cref{eq:lossFunction} using the standard back-propagation, as the whole system is fully differentiable.

Similarly to~\cite{monodepthv2}, the loss function consists of a photometric loss term $E_{\mbox{\scriptsize p}}$, which is a combination combination of $L^1$ and SSIM~\cite{wang2004image,lossfunctions} loss, and a disparity smoothness loss $E_{\mbox{\scriptsize dis}}$~\cite{monodepth17}.
The photometric loss $E_{\mbox{\scriptsize p}}$ also incorporates the auto-masking term $\mu_{ij}$\cite{monodepthv2}, which reduces the error when a patch becomes occluded or where a patch is no longer visible due to camera motion

\begin{align}
E_{\mbox{\scriptsize p}}(I^t, \hat{I}^{t}) &= \frac{1}{N}\sum_{i,j}\mu_{ij}P(I^t_{ij}, \hat{I}^{t}_{ij}) \\
P(p,q) &= \beta \frac{1-\mbox{SSIM}(p, q)}{2} + (1-\beta)\;\left\lVert p-q\right\rVert^1 \\
\mu_{ij} &= [P(I^t_{ij}, \hat{I}^{t}_{ij}) < P(I^t_{ij},I^{t+1}_{ij})]
\end{align}
where $[\cdot]$ indicates the Iverson bracket and consistently with the literature we set $\alpha=1$ and $\beta=0.85$.

\subsection{Instance Adaptation at Inference}\label{s:adapt}

We now introduce our main idea.
Thanks to the \textit{self-supervised formulation}, the loss function~\cref{eq:lossFunction} does not depend on any ground truth, but only on the observed images, and can thus be used at \emph{test time} to further improve the model by minimizing the same loss function through standard back-propagation for a defined number of steps.
As a result, we get a depth estimate which is optimized for each specific observation instance (pair of images) individually.

Note that this \textit{does not} imply training on testing data in the usual supervised learning sense, which would of course be invalid;
the reason is that there is no ground truth or additional supervision involved, but rather we simply perform an additional post-processing step at inference time for each sample. This idea is inspired by 3D human pose literature~\cite{joo2020exemplar}, where a 3D model is fitted to 2D detections at run time.

Generalizing, we explore two basic scenarios: In the first scenario (\textbf{instance-wise}), we process each temporarily adjacent pair of images individually without any temporal consistency between different pairs, always starting from the same model weights obtained during the training process of the baseline model.

In the second scenario (\textbf{sequential}), we exploit the fact video frames have spatial and temporal consistency. We start the inference for the first image in a sequence with the trained weights as in the previous case, but we then keep updating the model weights through out the whole video. This enables the model to aggregate information across multiple frames and it should also be faster as we shouldn't have to make as many updates for one individual frame as in the previous case. This scenario also better reflects practical deployment in a moving vehicle, where frames are processed in a stream as they are captured, rather than independently in an undefined order.

\subsection{Components and Parameterization}\label{s:components}

Note that the method of~\cref{s:adapt} can also be interpreted as learning more information from a single image pair, or a single video sequence.
An obvious concern is whether this might lead to \emph{over-fitting} and poor performance.
We found empirically that this is indeed the case --- our method works well, but only provided that the adaptation is limited to the correct subset of model parameters (see \cref{tab:interventions}).

Concretely, consider the current state-of-the-art monocular depth estimation pipeline~\cite{monodepthv2}, shown in~\Cref{fig:architecture}.
This model comprises four sub-networks: a depth encoder, a pose encoder, a depth decoder and a pose decoder.
When minimizing~\cref{eq:lossFunction}, we can optimize all sub-networks.
However, the depth and pose decoders can also be interpreted as \emph{parameterization} of depth and pose, expressing them as a function of corresponding codes computed by the encoders.
We find that updating the encoders while keeping the decoders fixed works the best because the system can adapt to the visual appearance of the novel image, whilst keeping the prior knowledge about different object types static in the decoders.
This and several other modelling choices are thoroughly assessed in~\cref{s:e.ablation}.

\input{tables/Quant-results}

%% file: tables/interventions.tex
\begin{table*}[t]
     \caption{The impact of enabling instance adaption for different network modules on the KITTI dataset\cite{Geiger2012CVPR}. For every configuration, we ran 10 SGD steps with LR=$10^-1$ and BatchNorm updates disabled.}\label{tab:interventions}
  \centering
  \vspace{-10pt}
  \resizebox{0.9\textwidth}{!}{
  \begin{tabular}{|c|c|c|c||c|c|c|c|c|c|c|c|}
  \hline
  Depth Encoder & Depth Decoder & Pose Encoder & Pose Decoder  & \cellcolor{col1}Abs Rel & \cellcolor{col1}Sq Rel & \cellcolor{col1}RMSE  & \cellcolor{col1}RMSE log & \cellcolor{col2}$\delta < 1.25 $ & \cellcolor{col2}$\delta < 1.25^{2}$ & \cellcolor{col2}$\delta < 1.25^{3}$ \\
  \hline

 \multicolumn{4}{|c||}{baseline\cite{monodepthv2} (no instance adaptation)} &  0.115 & 0.898 & 4.728 & 0.190 & 0.879 & 0.961 & 0.982 \\

 \arrayrulecolor{grey}\hline

\multicolumn{4}{|c||}{\cellcolor{green}whole network}
 & 0.207 & 3.152 & 8.008 & 0.272 & 0.729 & 0.891 & 0.951  \\

\arrayrulecolor{grey}\hline

\cellcolor{green}\checkmark& & & & 0.099 & 0.797 & 4.513 & 0.179 & 0.901 & 0.963 & 0.982  \\
\arrayrulecolor{grey}\hline
&\cellcolor{green}\checkmark & & & 0.156 & 1.572 & 6.379 & 0.235 & 0.797 & 0.932 & 0.972  \\
\arrayrulecolor{grey}\hline
& &\cellcolor{green}\checkmark & & 0.115 & 0.905 & 4.863 & 0.193 & 0.877 & 0.959 & 0.981  \\
\arrayrulecolor{grey}\hline
& & &\cellcolor{green}\checkmark & 0.115 & 0.905 & 4.863 & 0.193 & 0.877 & 0.959 & 0.981  \\
\arrayrulecolor{grey}\hline

\cellcolor{green}\checkmark& \cellcolor{green}\checkmark & & & 0.156 & 1.58 & 6.463 & 0.235 & 0.795 & 0.931 & 0.971  \\
\arrayrulecolor{grey}\hline
\cellcolor{green}\checkmark& &\cellcolor{green}\checkmark & & \textbf{0.097} & \textbf{0.788} & \textbf{4.464} & \textbf{0.178} & \textbf{0.905} &\textbf{ 0.964} & \textbf{0.982}  \\
\arrayrulecolor{grey}\hline
\cellcolor{green}\checkmark & & & \cellcolor{green}\checkmark& 0.111 & 0.895 & 4.776 & 0.19 & 0.884 & 0.959 & 0.981  \\
\arrayrulecolor{grey}\hline
& \cellcolor{green}\checkmark &\cellcolor{green}\checkmark & & 0.151 & 1.443 & 6.217 & 0.229 & 0.806 & 0.936 & 0.973  \\
\arrayrulecolor{grey}\hline
& \cellcolor{green}\checkmark& & \cellcolor{green}\checkmark& 0.208 & 3.143 & 8.082 & 0.273 & 0.728 & 0.892 & 0.952  \\
\arrayrulecolor{grey}\hline
& & \cellcolor{green}\checkmark&\cellcolor{green}\checkmark & 0.115 & 0.905 & 4.863 & 0.193 & 0.877 & 0.959 & 0.981  \\
\arrayrulecolor{grey}\hline

\cellcolor{green}\checkmark&\cellcolor{green}\checkmark &\cellcolor{green}\checkmark & & 0.113 & 0.912 & 4.804 & 0.191 & 0.881 & 0.958 & 0.981  \\
\arrayrulecolor{grey}\hline
\cellcolor{green}\checkmark&\cellcolor{green}\checkmark& &\cellcolor{green}\checkmark & 0.212 & 3.319 & 8.183 & 0.276 & 0.725 & 0.887 & 0.948 \\
\arrayrulecolor{grey}\hline
\cellcolor{green}\checkmark& &\cellcolor{green}\checkmark & \cellcolor{green}\checkmark& 0.154 & 1.561 & 6.374 & 0.233 & 0.802 & 0.932 & 0.971  \\
\arrayrulecolor{grey}\hline
&\cellcolor{green}\checkmark &\cellcolor{green}\checkmark &\cellcolor{green}\checkmark & 0.199 & 2.927 & 7.811 & 0.266 & 0.741 & 0.898 & 0.955 \\

\arrayrulecolor{grey}\hline

\cellcolor{lightgreen}first layer& & & & 0.111 & 0.884 & 4.811 & 0.191 & 0.881 & 0.959 & 0.981 \\
\arrayrulecolor{grey}\hline
& & \cellcolor{lightgreen}first layer & & 0.115 & 0.905 & 4.863 & 0.193 & 0.877 & 0.959 & 0.981 \\
\arrayrulecolor{grey}\hline
\cellcolor{lightgreen}last residual block & & & & 0.104 & 0.828 & 4.64 & 0.184 & 0.893 & 0.962 & 0.982  \\
\arrayrulecolor{grey}\hline
& & \cellcolor{lightgreen}last residual block & & 0.115 & 0.905 & 4.863 & 0.193 & 0.877 & 0.959 & 0.981  \\
\arrayrulecolor{grey}\hline
\cellcolor{lightgreen}first layer & & \cellcolor{lightgreen}first layer & & 0.11 & 0.878 & 4.797 & 0.19 & 0.883 & 0.959 & 0.981  \\
\arrayrulecolor{grey}\hline
\cellcolor{lightgreen}last residual block & & \cellcolor{lightgreen}last residual block & & 0.103 & 0.821 & 4.622 & 0.183 & 0.894 & 0.963 & 0.982  \\
\arrayrulecolor{grey}\hline

\arrayrulecolor{black}\hline

  \end{tabular}
  }
 \vspace{-15pt}

\end{table*}

%% file: tables/Quant-results.tex
\begin{table*}[t!]
  \setlength{\tabcolsep}{3pt}
  \caption{Depth estimation accuracy on the KITTI dataset~\cite{Geiger2012CVPR} using the Eigen split~\cite{EigenF14}}
  \centering
  \begin{minipage}[t]{0.80\textwidth}
  \vspace{-10pt}
  \footnotesize{
  \begin{tabular}{|l|c||c|c|c|c|c|c|c|}
  \hline
  Method & Train & \cellcolor{col1}Abs Rel & \cellcolor{col1}Sq Rel & \cellcolor{col1}RMSE  & \cellcolor{col1}RMSE log & \cellcolor{col2}$\delta<1.25 $ & \cellcolor{col2}$\delta<1.25^{2}$ & \cellcolor{col2}$\delta<1.25^{3}$\\
  \hline
Eigen~\cite{EigenF14} & D & 0.203 & 1.548 & 6.307 & 0.282 & 0.702 & 0.890 & 0.890\\
Liu~\cite{LiuSLR15} & D & 0.201 & 1.584 & 6.471 & 0.273 & 0.680 & 0.898 & 0.967\\
Klodt~\cite{klodt2018supervising} & D*M & 0.166 & 1.490 & 5.998 & --- &  0.778 & 0.919 & 0.966\\
AdaDepth~\cite{gandepth2018}  & D* & 0.167 & 1.257 & 5.578 & 0.237 & 0.771 & 0.922 & 0.971\\
Kuznietsov~\cite{KuznietsovSL17} & DS & 0.113 & 0.741 & 4.621 & 0.189 & 0.862 & 0.960 & 0.986\\
DVSO~\cite{yang2018deep} & D*S & 0.097 & 0.734 & 4.442 & 0.187 & 0.888 & 0.958 & 0.980\\
SVSM FT~\cite{singlestereo2018} & DS & \underline{0.094} & \underline{0.626} & 4.252 & 0.177 & 0.891 & 0.965 & 0.984\\
Guo~\cite{guo2018learning} & DS & 0.096 & 0.641  & \underline{4.095}  & \underline{0.168}  & \underline{0.892}  & \underline{0.967}  & \underline{0.986} \\
DORN~\cite{Fu_2018_CVPR} & D & \textbf{0.072}&  \textbf{0.307} & \textbf{2.727} & \textbf{0.120} & \textbf{0.932} & \textbf{0.984} & \textbf{0.994}\\

\arrayrulecolor{black}\hline

Zhou~\cite{zhou2017}\textdagger & M & 0.183 & 1.595 & 6.709 & 0.270 & 0.734 & 0.902 & 0.959\\
Yang~\cite{yang2017unsupervised} & M & 0.182 & 1.481  & 6.501  & 0.267  & 0.725  & 0.906  & 0.963\\
Mahjourian~\cite{mahjourian2018unsupervised} & M & 0.163 & 1.240 & 6.220 & 0.250 & 0.762 & 0.916 & 0.968\\
GeoNet~\cite{Yin18}\textdagger & M  & 0.149 & 1.060 & 5.567 & 0.226 & 0.796 & 0.935 & 0.975\\
DDVO~\cite{wang2017learning} & M  & 0.151 & 1.257 & 5.583 & 0.228 & 0.810 & 0.936 & 0.974\\
DF-Net~\cite{zou2018df} & M & 0.150 & 1.124 & 5.507 & 0.223 & 0.806 & 0.933 & 0.973\\
LEGO~\cite{yang2018lego} & M & 0.162 & 1.352 & 6.276 & 0.252 & --- & --- & --- \\
Ranjan~\cite{ranjan2018adversarial}  & M & 0.148 & 1.149 & 5.464 & 0.226 & 0.815 & 0.935 & 0.973\\
EPC++~\cite{luo2018every} & M & 0.141 & 1.029 & 5.350 & 0.216 & 0.816 & 0.941 & 0.976\\
Struct2depth `(M)'~\cite{casser2018depth}  & M & 0.141 & 1.026 & 5.291 &  0.215 & 0.816 & 0.945 & 0.979\\
Monodepth~V2\cite{monodepthv2} & M &      0.115 &      0.903 &      4.863 &      0.193 &      0.877 &      0.959 &      0.981 \\
Monodepth~V2\cite{monodepthv2} (1024 $\times$ 320)  &  M &   0.115&     0.882&     4.701&     0.190&     0.879&     0.961&     0.982 \\
Patil et al.~\cite{patil2020dont} & M& \underline{0.111} & \underline{0.821} & \underline{4.65} & \underline{0.187} & \underline{0.883} & \underline{0.961} & \underline{0.982}  \\

\arrayrulecolor{grey}
\textit{ours} & M & {\bf 0.089} & {\bf 0.747} & {\bf 4.275} & 0.173 & 0.912 & 0.964 & 0.982 \\

\arrayrulecolor{grey}
\textit{ours} (1024$\times$320) & M & 0.089 & 0.756 &  4.228 &  \textbf{0.170} &  \textbf{0.917} &  \textbf{0.967} &  \textbf{0.983} \\

\arrayrulecolor{black}\hline

Garg~\cite{GargBR16}\textdagger & S  &  0.152 & 1.226 & 5.849 & 0.246 & 0.784 & 0.921 & 0.967\\
Monodepth R50~\cite{monodepth17}\textdagger & S & 0.133 & 1.142 & 5.533 & 0.230 & 0.830 & 0.936 & 0.970\\
StrAT~\cite{mehta2018structured}  & S  &  0.128 & 1.019 & 5.403 & 0.227 & 0.827 & 0.935 & 0.971\\
3Net  (R50)~\cite{poggi20183net} & S & 0.129 & 0.996 & 5.281 & 0.223 & 0.831 & 0.939 & 0.974 \\
3Net (VGG)~\cite{poggi20183net} & S & 0.119 &  1.201 & 5.888 & 0.208 &  0.844 & 0.941 & 0.978  \\
SuperDepth+pp~\cite{pillai2018superdepth} (1024$\times$382)  & S & 0.112 & 0.875 & 4.958 & 0.207 & 0.852 & 0.947 & \underline{0.977} \\
Monodepth~V2\cite{monodepthv2}      &  S  &  0.109  &   0.873 &   4.960 &   0.209 &     0.864&     0.948 &   0.975 \\
Monodepth~V2\cite{monodepthv2} (1024 $\times$ 320) &   S &  0.107 &     0.849&     4.764&     0.201&     0.874&     0.953 &   \underline{0.977} \\

\arrayrulecolor{grey}

\textit{ours} & S & 0.076 & 0.691 & 4.264 & 0.182 & 0.916 & 0.958 & 0.976 \\

\arrayrulecolor{grey}
\textit{ours} (1024$\times$320) & S & \textbf{0.075} & \textbf{0.683} & \textbf{4.186} & \textbf{0.179} & \textbf{0.918} &\textbf{ 0.960} & \textbf{0.978} \\

\arrayrulecolor{black}\hline

UnDeepVO~\cite{li2017undeepvo} & MS  &  0.183 & 1.730 & 6.57 & 0.268 & --- & --- & -\\
Zhan~FullNYU~\cite{zhanst2018} & D*MS  &  0.135 & 1.132 & 5.585 & 0.229 & 0.820 & 0.933 & 0.971\\
EPC++~\cite{luo2018every} & MS & 0.128 & \underline{0.935} & \underline{5.011} & \underline{0.209} & 0.831 & \underline{0.945} & 0.979 \\
monodepth~\cite{monodepthv2}& MS &
   0.106&     0.818&     4.750&     0.196&     0.874&     0.957&     0.979 \\

monodepth~\cite{monodepthv2} (1024 $\times$ 320) & MS
&   0.106 &     0.806&     4.630&     0.193&     0.876&     0.958&     0.980  \\

\arrayrulecolor{grey}
\textit{ours} & MS & {\bf 0.074} & {\bf 0.636} & 4.082 &  0.168 &  0.924 &  0.966 &  0.982 \\

\arrayrulecolor{grey}
\textit{ours} (1024$\times$320) & MS & 0.076 & 0.638 & {\bf 4.044} & {\bf 0.166} & {\bf 0.925} & {\bf 0.967} & {\bf 0.982} \\

\arrayrulecolor{black}\hline

\end{tabular}}
\end{minipage}%
\begin{minipage}[t]{0.20\textwidth}
\vspace{110pt}
{\scriptsize
\textbf{Legend} \newline
D  --- Depth supervision \newline
D* --- Auxiliary depth supervision \newline
S  --- Self-supervised stereo \newline
M  --- Self-supervised mono \newline
\textdagger~--- Newer results from GitHub \newline
+ pp --- With post-processing \newline
\newline
For {\color{col1} \textbf{red}} metrics, the lower is better; for {\color{col2} \textbf{blue}} metrics, the higher is better. \newline \newline
Best results in each category are in \textbf{bold}; second-best \underline{underlined}}
\end{minipage}\label{tab:kitti_eigen}
\vspace{-5pt}
\end{table*}

%% file: experiments.tex
\section{Experiments}
\input{tables/LR}
\input{tables/epochs}
\input{tables/opt-outputs}

After discussing the experimental data (\cref{s:e.data}), we first optimize and ablate the various component of our approach (\cref{s:e.ablation}), followed by a thorough evaluation of the the instance- and sequential-wise scenarios (\cref{s:e.instance,s:e.sequence}).

\subsection{Data}\label{s:e.data}

We evaluate our method using the standard KITTI 2015 Stereo dataset\cite{Geiger2012CVPR}.
Consistently with the literature\cite{monodepth17, monodepthv2,casser2018depth, luo2018every}, we use the data split of Eigen et al.\cite{EigenF14}, where for monocular setup we follow the pre-processing of Zhou et al.~\cite{zhou2017}, which removes stationary scenes which would otherwise prove problematic for image warping based methods.
This results in $39810$ images with temporally adjacent frames for training, $4424$ for validation and $697$ for testing.
In stereo and mixed monocular stereo set up, we use the full Eigen dataset, again consistently with the prior work. For monocular case, the method output is rescaled using the per-image median ground truth scaling as in~\cite{zhou2017, monodepthv2}; for the stereo case, absolute depth can be inferred directly using the camera baseline as a scaling factor.

\subsection{Optimal Configuration and Ablation}\label{s:e.ablation}

In order to find the optimal setting for our method, we validate which model components to adapt (\cref{s:e.components}), how fast (\cref{s:e.lr}) and for how long (\cref{s:e.iter}).
We also compare our method to a simple baseline in which depth and pose are optimized directly (\cref{s:e.direct}), which can also be interpreted as a special case of our approach.

\subsubsection{Components}\label{s:e.components}

As noted in~\cref{s:components}, the success of our method depends strongly on which parts of the model are optimized.
This is assessed in~\cref{tab:interventions}.
As seen in the second row, simply updating all network parameters at inference time makes the accuracy significantly worse than the baseline (first line).
We therefore systematically evaluate which sections of the network can have their weights updated without causing divergence.
As described in~\cref{s:components}, our network can be split into four modules: two encoders based on a ResNet-18~\cite{he2016deep} architecture and two decoders --- one which takes skip connections from the depth encoder to generate a depth prediction and the pose decoder.
We also disable updates of all Batch Normalization~\cite{ioffe2015batch} layers, as we found that updating BatchNorm parameters during inference also causes divergence.
We first evaluate each segment separately, then all possible combinations and finally subsections of the most successful network sub-components (see \cref{tab:interventions}).
We conclude that if only weights of the pose and the depth encoders are updated at inference, the depth estimation accuracy improves the most.

\subsubsection{Learning Rate}\label{s:e.lr}

At training time, the model is trained with Adam~\cite{kingma2014adam} with learning rate $10^{-4}$, consistently with~\cite{monodepthv2}. Adam however requires warm-up iterations, so at inference time we use vanilla SGD and therefore choosing the correct learning rate to update model parameters is an important consideration, as too low rate results in little to no change, but too high learning rate results in quick divergence.
In \cref{tab:LR} we explore a range of reasonable learning rates by taking 10 updates steps and observing the overall accuracy, and we conclude that the learning rate of $0.1$ is the best as it results in the most significant improvement in results.

\subsubsection{Number of Optimization Steps}\label{s:e.iter}

The next natural question is how long we can update network parameters to improve performance while being time-efficient and not causing divergence. \Cref{tab:epochs} shows that with learning rate $0.1$ we can see the that even a low number of steps allows us to still benefit and therefore taking 50 steps results in a good balance between time efficiency and performance improvement.

\input{tables/odom_seq}

\subsubsection{Direct Optimization}\label{s:e.direct}

Instead of optimizing the weights of the network, we can also consider a ``naive'' approach by optimizing the networks outputs directly.
Namely, we take the depth map and pose outputs and optimise them by taking the loss function \cref{eq:lossFunction} and using the gradients to update the depth map and the pose estimate directly.
As we show in~\cref{tab:opt-outputs}, this naive approach does not improve the baseline.
Optimizing only the depth map and not the pose prediction performed similarly poor (results not shown for brevity).

\subsection{Instance-wise Inference}\label{s:e.instance}

In the instance-wise strategy,  weights are modified individually for each image, without any information from previously observed frames. Resetting weights after adapting to an instance can be beneficial as it removes the possibility of overfitting a sequence very different to generally observed scenes, but may result in lack of temporal consistency. In this experiment, for each image we start by taking the weights obtained by training the baseline method\cite{monodepthv2} on the training split and we take $50$ steps of SGD with learning rate $0.1$. Our method consistently beats all the previous methods (see \Cref{tab:kitti_eigen}) by a significant margin. We also present results in higher resolution as~\cite{pillai2018superdepth, monodepthv2} show this can result in greater performance, and for our method this in some cases improves accuracy even further.

Using our method, we can also continue to learn on instances of stereo image pairs --- in this case we only optimize the depth encoder as pose differences between images are from a known camera baseline. We again consistently outperform all the previous results by a significant margin, and despite being a self-supervised method trained without any ground truth, we come close to DORN~\cite{Fu_2018_CVPR}, the the state-of-the-art \textit{fully-supervised} depth estimation method.

Combining stereo and monocular data in training is also a popular strategy employed in~\cite{li2017undeepvo, monodepthv2,luo2018every}.  In this case we warp the stereo pairs using the known baseline and a pose network is used for temporally adjacent pairs, the method therefore utilizes both monocular temporally adjacent frames and stereo pairs in training and instance-wise adaption. Our method also shows significant improvement, demonstrating  robustness of our method to improve with various types of instance data (see \Cref{tab:kitti_eigen} --- bottom section).

\begin{figure*}[b]
    \centering
    \input{figs/seq-graph}
    \vspace{-20pt}
    \caption{Absolute depth estimation error for individual frames in the KITTI Odometry 9 sequence}\label{fig:sequence}
\end{figure*}
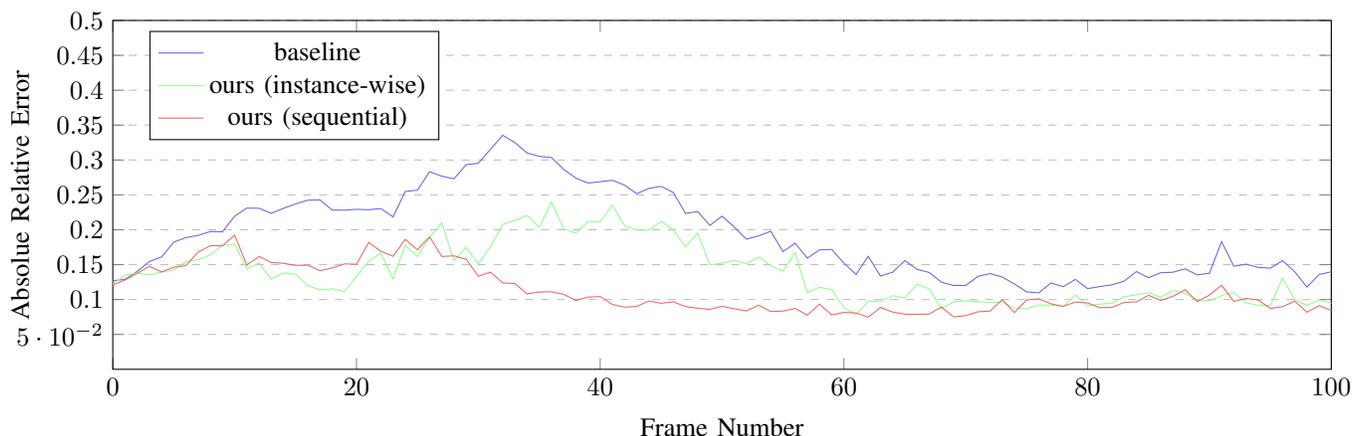

\subsection{Sequential Inference}\label{s:e.sequence}

In the sequential strategy, we again start by taking the weights obtained by training the baseline method\cite{monodepthv2} on the training split for the first frame in the video sequence, but then we keep updating the weights using only 5 steps for each subsequent frame throughout the whole sequence. We again evaluate on the Eigen split\cite{EigenF14}, but because the split contains only short sequences and we want to test robustness of this approach for long sequences, we also evaluate on the KITTI Odometry test split\cite{Geiger2012CVPR}, which contains 11 sequences between 491 and 4981 frames in length.

We observe that for both sets, the sequential strategy also consistently improves over the baseline, performing approximately at par with the instance-wise strategy, but \textit{being 10 times faster} (see \cref{tab:odom}). The experiment also shows that even for long video sequences, the model still performs well and does not diverge (see \cref{fig:sequence}), despite the fact it had already made thousands of self-supervised updates to the baseline model (e.g.~for a video of 4481 frames, the method makes $4980\times 5=24,900$ updates, so one needs to ask whether the model parameters wouldn't ``drift away'' after so many steps).



%% file: tables/LR.tex
\begin{table}
  \centering
  \caption{Instance adaptation learning rate ablation }
  \vspace{-10pt}
  \resizebox{\textwidth}{!}{
  \begin{tabular}{|c|c|c|c|c|c|c|c|c|c|}
  \hline
  Learning Rate  & \cellcolor{col1}Abs Rel & \cellcolor{col1}Sq Rel & \cellcolor{col1}RMSE  & \cellcolor{col1}RMSE log & \cellcolor{col2}$\delta < 1.25 $ & \cellcolor{col2}$\delta < 1.25^{2}$ & \cellcolor{col2}$\delta < 1.25^{3}$\\
  
  \hline
  
baseline & 0.115 & 0.905 & 4.863 & 0.193 & 0.877 & 0.959 & 0.981 \\
\arrayrulecolor{grey}\hline
$10^{-5}$ & 0.115 & 0.905 & 4.863 & 0.193 & 0.877 & 0.959 & 0.981 \\
$10^{-4}$ & 0.114 & 0.904 & 4.861 & 0.193 & 0.878 & 0.959 & 0.981 \\
$10^{-3}$ & 0.113 & 0.897 & 4.842 & 0.192 & 0.879 & 0.96 & 0.981 \\
$10^{-2}$ & 0.107 & 0.852 & 4.719 & 0.186 & 0.889 & 0.961 & 0.981 \\
$10^{-1}$ & 0.097 & 0.774 & 4.455 & 0.178 & 0.904 & 0.964 & 0.982 \\
1 & 0.279 & 2.763 & 8.577 & 0.384 & 0.563 & 0.78 & 0.894 \\
10 & 0.451 & 5.596 & 12.153 & 0.591 & 0.314 & 0.57 & 0.768 \\

\arrayrulecolor{black}\hline

  \end{tabular}
  }
\label{tab:LR}  
\vspace{-10pt}
\end{table}

%% file: tables/epochs.tex
\begin{table}[!t]
  \centering
  \caption{Number of optimization steps ablation}
  \vspace{-10pt}
  \resizebox{\textwidth}{!}{
  \begin{tabular}{|c|c|c|c|c|c|c|c|c|c|}
  \hline
  Steps  & \cellcolor{col1}Abs Rel & \cellcolor{col1}Sq Rel & \cellcolor{col1}RMSE  & \cellcolor{col1}RMSE log & \cellcolor{col2}$\delta < 1.25 $ & \cellcolor{col2}$\delta < 1.25^{2}$ & \cellcolor{col2}$\delta < 1.25^{3}$ &  \cellcolor{col1}Seconds per Image\\
  
  \hline
Initial & 0.115 & 0.905 & 4.863 & 0.193 & 0.877 & 0.959 & 0.981 & 0.022 \\
\arrayrulecolor{grey}\hline
5 & 0.102 & 0.806 & 4.566 & 0.182 & 0.897 & 0.962 & 0.982 & 0.291 \\
10 & 0.097 & 0.788 & 4.464 & 0.178 & 0.905 & 0.964 & 0.982 & 0.544 \\
25 & 0.091 & 0.752 & 4.331 & 0.174 & 0.911 & 0.964 & 0.982 & 1.21 \\
50 & 0.089 & 0.742 & 4.263 & 0.172 & 0.913 & 0.964 & 0.982 & 2.354 \\
75 & 0.088 & 0.735 & 4.226 & 0.171 & 0.914 & 0.965 & 0.982 & 3.551 \\
100 & 0.086 & 0.723 & 4.22 & 0.17 & 0.916 & 0.965 & 0.982 & 4.626 \\
150 & 0.089 & 0.759 & 4.219 & 0.174 & 0.91 & 0.962 & 0.98 & 6.943 \\
200 & 0.09 & 0.786 & 4.241 & 0.176 & 0.908 & 0.961 & 0.98 & 9.355 \\

\arrayrulecolor{black}\hline

  \end{tabular}
  }
\label{tab:epochs}  
\vspace{-10pt}
\end{table}

%% file: tables/opt-outputs.tex
\begin{table}
\caption{Ablation: direct depth optimization}
\vspace{-10pt}
\centering
\resizebox{\textwidth}{!}{
\begin{tabular}{|c|c|c|c|c|c|c|c|c|c|}
\hline
Learning Rate & \cellcolor{col1}Abs Rel & \cellcolor{col1}Sq Rel & \cellcolor{col1}RMSE  & \cellcolor{col1}RMSE log & \cellcolor{col2}$\delta < 1.25 $ & \cellcolor{col2}$\delta < 1.25^{2}$ & \cellcolor{col2}$\delta < 1.25^{3}$ \\

\hline
$10^{-4}$ & 0.115 & 0.885 & 4.705 & 0.19 & 0.879 & 0.961 & 0.982  \\
$10^{-2}$ & 0.115 & 0.885 & 4.705 & 0.19 & 0.879 & 0.961 & 0.982  \\
1 & 0.115 & 0.884 & 4.7 & 0.19 & 0.879 & 0.961 & 0.982  \\
10 & 0.114 & 0.878 & 4.677 & 0.189 & 0.88 & 0.962 & 0.982  \\

\arrayrulecolor{black}\hline
\end{tabular}
}\label{tab:opt-outputs}\hfill
 
\end{table}

%% file: tables/odom_seq.tex
\begin{table*}
\caption{Qualitative results on the KITTI test set. The Odometry split consist of image sequences of approx. 1000 frames}
\vspace{-10pt}
\footnotesize{
\centering
\begin{tabular}{|l|l|c|c|c|c|c|c|c|c|c|c|}
\hline
data split & strategy  & Steps &  LR & \cellcolor{col1}Abs Rel & \cellcolor{col1}Sq Rel & \cellcolor{col1}RMSE  & \cellcolor{col1}RMSE log & \cellcolor{col2}$\delta < 1.25 $ & \cellcolor{col2}$\delta < 1.25^{2}$ & \cellcolor{col2}$\delta < 1.25^{3}$ & \cellcolor{col1} time [$s$] \\ \hline
Eigen\cite{EigenF14} & baseline\cite{monodepthv2} & --- &  --- &  0.115 &      0.903 &      4.863 &      0.193 &      0.877 &      0.959 &      0.981 & 0.022\\
 & Patil et al.~\cite{patil2020dont}&  --- &  --- &  0.111 & 0.821 & 4.65 & 0.187 & 0.883 & 0.961 & 0.982 & --- \\
 & ours (instance-wise) & 50 & 0.1 &  0.089 &  0.747 &  4.275 &  0.173 &  0.912 &  0.964 &  0.982 & 2.354\\

 & ours (sequence-wise) &  5 & 0.1 &  0.105  &  0.896  &  4.707  &  0.186  &  0.895  &  0.960  &  0.981 & 0.291 \\

\hline


Odometry & baseline\cite{monodepthv2} &  --- &  --- & 0.147 & 1.090 & 5.171 & 0.219 & 0.805 & 0.942 & 0.978 & 0.022\\
& ours (instance-wise) & 50 & 0.01 & 0.116 & 0.895 & 4.289 & 0.181 & 0.772 & 0.867 & 0.893 & 2.354 \\
& ours (sequence-wise) & 5 & 0.01 & 0.113 & 0.887 & 4.454 & 0.186 & 0.874 & 0.956 & 0.981 & 0.291\\




\arrayrulecolor{black}\hline

\end{tabular}
}\label{tab:odom}\vspace{-10pt}
\end{table*}

%% file: figs/seq-graph.tex
\begin{tikzpicture}
\begin{axis}[
    xlabel={Frame Number},
    ylabel={Absolue Relative Error},
    xmin=0, xmax=100,
    ymin=0, ymax=0.5,
    xtick={0,20,40,60,80,100},
    ytick={0.05, 0.1, 0.15, 0.2, 0.25, 0.3, 0.35, 0.4, 0.45, 0.5},
    legend pos=north west,
    ymajorgrids=true,
    grid style=dashed,
    width=\textwidth,
    height=0.35\textwidth,
]

\addplot[color=blue]
    coordinates {

    (0,0.12697338)(1,0.12883686)(2,0.14032845)(3,0.15447155)(4,0.1610449)(5,0.18229826)(6,0.18875474)(7,0.19191518)(8,0.19743915)(9,0.19700548)(10,0.21950385)(11,0.2310965)(12,0.2307723)(13,0.22360244)(14,0.23066919)(15,0.23699571)(16,0.24238893)(17,0.24272949)(18,0.22830446)(19,0.22802983)(20,0.22930086)(21,0.2287098)(22,0.2302958)(23,0.21844082)(24,0.25494796)(25,0.25666937)(26,0.28321174)(27,0.27675778)(28,0.2730351)(29,0.29332313)(30,0.29531437)(31,0.31539583)(32,0.3354069)(33,0.32486907)(34,0.31002054)(35,0.30523953)(36,0.3034609)(37,0.28662276)(38,0.27408928)(39,0.26694262)(40,0.26886082)(41,0.27093196)(42,0.26394287)(43,0.25169495)(44,0.25931296)(45,0.2621826)(46,0.2534051)(47,0.22348362)(48,0.22602265)(49,0.20631145)(50,0.2195071)(51,0.20394543)(52,0.18652613)(53,0.19115756)(54,0.19784597)(55,0.16871785)(56,0.1806587)(57,0.15928935)(58,0.17120722)(59,0.17180146)(60,0.15226208)(61,0.13602468)(62,0.16188155)(63,0.13364583)(64,0.13909173)(65,0.15578988)(66,0.14331476)(67,0.13874833)(68,0.12492315)(69,0.12017707)(70,0.120142035)(71,0.13311836)(72,0.13721661)(73,0.13219523)(74,0.1222585)(75,0.11101008)(76,0.10961677)(77,0.1239263)(78,0.1183868)(79,0.12880763)(80,0.11565531)(81,0.11856491)(82,0.120990366)(83,0.12641576)(84,0.14010087)(85,0.13122627)(86,0.13844632)(87,0.13911058)(88,0.1439636)(89,0.13535129)(90,0.13769493)(91,0.18289585)(92,0.14777394)(93,0.15073378)(94,0.14597815)(95,0.14500277)(96,0.15574917)(97,0.13981196)(98,0.11780783)(99,0.13585265)(100,0.13975738)
    };

\addplot[color=green]
    coordinates {
    (0,0.12214379)(1,0.13474661)(2,0.13813403)(3,0.13501604)(4,0.13927254)(5,0.14235575)(6,0.15421931)(7,0.15703182)(8,0.16370888)(9,0.1777458)(10,0.1794432)(11,0.14385122)(12,0.15252711)(13,0.12930436)(14,0.1381045)(15,0.13648772)(16,0.119941376)(17,0.11389077)(18,0.1154551)(19,0.11129043)(20,0.13305014)(21,0.15500653)(22,0.16628721)(23,0.12961175)(24,0.17759846)(25,0.16169131)(26,0.18913002)(27,0.21006027)(28,0.1558393)(29,0.1748623)(30,0.15121971)(31,0.1753644)(32,0.20787491)(33,0.21358946)(34,0.22026268)(35,0.20355578)(36,0.23982477)(37,0.2013575)(38,0.19517682)(39,0.2119679)(40,0.21090643)(41,0.2356493)(42,0.2057121)(43,0.1993451)(44,0.19958661)(45,0.21215092)(46,0.20009416)(47,0.17549694)(48,0.19501722)(49,0.14977475)(50,0.1519458)(51,0.15608674)(52,0.15140486)(53,0.16085668)(54,0.14841942)(55,0.14066991)(56,0.16707036)(57,0.11011257)(58,0.11768365)(59,0.11413122)(60,0.08839302)(61,0.07838809)(62,0.09741393)(63,0.09763414)(64,0.105537675)(65,0.10229851)(66,0.12208258)(67,0.11536294)(68,0.087037355)(69,0.0966382)(70,0.09804445)(71,0.097126976)(72,0.09553415)(73,0.09631543)(74,0.0875706)(75,0.086683504)(76,0.09211552)(77,0.09210223)(78,0.08977344)(79,0.106489204)(80,0.09070866)(81,0.09359862)(82,0.09444601)(83,0.104133464)(84,0.106705576)(85,0.10967151)(86,0.103068605)(87,0.112961456)(88,0.10910094)(89,0.0988213)(90,0.09821283)(91,0.10561557)(92,0.11027334)(93,0.09519688)(94,0.092041545)(95,0.09116651)(96,0.13068925)(97,0.1002601)(98,0.091547795)(99,0.09941129)(100,0.09484323) 
    };

\addplot[color=red]
    coordinates {
    (0,0.12078845)(1,0.12793016)(2,0.13756594)(3,0.14756994)(4,0.13954376)(5,0.1462912)(6,0.14865619)(7,0.16794886)(8,0.17704254)(9,0.17711504)(10,0.19225225)(11,0.1494803)(12,0.16166517)(13,0.1530539)(14,0.15191865)(15,0.1489202)(16,0.14928871)(17,0.14132807)(18,0.14525577)(19,0.15100852)(20,0.15062222)(21,0.18178149)(22,0.16920616)(23,0.16188)(24,0.18619579)(25,0.1712354)(26,0.18981075)(27,0.16176368)(28,0.16256301)(29,0.15776904)(30,0.13344598)(31,0.13934405)(32,0.12383665)(33,0.12285372)(34,0.10820442)(35,0.1106989)(36,0.11119301)(37,0.10741614)(38,0.09871067)(39,0.10344998)(40,0.10426269)(41,0.092857085)(42,0.08902426)(43,0.09014383)(44,0.097975366)(45,0.09460009)(46,0.096746586)(47,0.08979905)(48,0.08754955)(49,0.085957244)(50,0.09036431)(51,0.08673742)(52,0.083751485)(53,0.091770515)(54,0.08285961)(55,0.08344377)(56,0.08725846)(57,0.077611685)(58,0.09334424)(59,0.07794753)(60,0.08159751)(61,0.080673225)(62,0.07468844)(63,0.08860228)(64,0.082120836)(65,0.07897931)(66,0.07881761)(67,0.079099625)(68,0.08934708)(69,0.07507492)(70,0.07700557)(71,0.08239364)(72,0.08354499)(73,0.09965301)(74,0.08126904)(75,0.099442266)(76,0.1010715)(77,0.09383113)(78,0.09030721)(79,0.09629402)(80,0.095128976)(81,0.08822063)(82,0.0888504)(83,0.09584441)(84,0.09664935)(85,0.106039874)(86,0.09876836)(87,0.10450746)(88,0.11398249)(89,0.09711835)(90,0.106352136)(91,0.12035973)(92,0.09733162)(93,0.10162872)(94,0.09930049)(95,0.08734449)(96,0.0896629)(97,0.097730905)(98,0.08185341)(99,0.091247104)(100,0.08431803)};

    \legend{baseline, ours (instance-wise), ours (sequential)}
    
\end{axis}
\end{tikzpicture}

%% file: conclusion.tex
\section{Conclusion}
In this paper, we introduced \textit{instance adaptation for self-supervised} depth estimation methods. The instance adaptation allows us to update certain parameters of a trained model dynamically during inference, improving the depth estimate for image instance individually.

Our self-supervised method consistently outperforms all the previous methods by a significant margin for all three common setups (monocular, stereo and monocular+stereo), and comes very close in accuracy when compared to the \emph{fully-supervised} state-of-the-art methods. Additionally, we presented the sequential improvement for our method, which is equally accurate but is 10 times faster than the instance-wise approach.